\documentclass[11pt,a4paper]{article}
\usepackage[hyperref]{character_relationships}
\usepackage{times}
\usepackage{latexsym}
\usepackage{graphicx}
\usepackage{amsmath}

\usepackage{url}

\aclfinalcopy 
\def\aclpaperid{25} 

\title{Character-Centric Storytelling}

\author{Aditya Surikuchi \\
  Aalto University \\
  \texttt{aditya.surikuchi@aalto.fi} \\\And
  Jorma Laaksonen \\
  Aalto University \\
  \texttt{jorma.laaksonen@aalto.fi} \\}

\date{}

\begin{document}
\maketitle
\begin{abstract}
  Sequential vision-to-language or visual storytelling has recently been one of the areas of focus in computer vision and language modeling domains. Though existing models generate narratives that read subjectively well, there could be cases when these models miss out on generating stories that account and address all prospective human and animal characters in the image sequences. Considering this scenario, we propose a model that implicitly learns relationships between provided characters and thereby generates stories with respective characters in scope. We use the VIST dataset for this purpose and report numerous statistics on the dataset. Eventually, we describe the model, explain the experiment and discuss our current status and future work.
\end{abstract}



\section{Introduction}

Visual storytelling and album summarization tasks have recently been of focus in the domain of computer vision and natural language processing. With the advent of new architectures, solutions for problems like image captioning and language modeling are getting better. Therefore it is only natural to work towards storytelling; deeper visual context yielding a more expressive style language, as it could potentially improve various applications involving tasks using visual descriptions and visual question answering. \cite{Wiriyathammabhum:2016:CVN:3022634.3009906}.

Since the release of the VIST visual storytelling dataset \cite{DBLP:journals/corr/HuangFMMADGHKBZ16}, there have been numerous approaches modeling the behavior of stories, leveraging and extending successful sequence-to-sequence based image captioning architectures. Some of them primarily addressed means of incorporating image-sequence feature information into a narrative generating network \cite{DBLP:journals/corr/abs-1806-00738}, \cite{DBLP:journals/corr/abs-1805-10973}, while others focused on model learning patterns and behavioral orientations with changes in back-propagation methods \cite{DBLP:journals/corr/abs-1804-09160}, \cite{DBLP:journals/corr/abs-1805-08191}. Motivated by these works we now want to understand the importance of characters and their relationships in visual storytelling.

Specifically, we extract characters from the VIST dataset, analyze their influence across the dataset and exploit them for paying attention to relevant visual segments during story-generation. We report our findings, discuss the directions of our ongoing work and suggest recommendations for using characters as semantics in visual storytelling.


\section{Related work}

\cite{DBLP:journals/corr/HuangFMMADGHKBZ16} published the VIST dataset along with a baseline sequence-to-sequence learning model that generates stories for image sequences in the dataset. Gradually, as a result of the 2018 storytelling challenge, there have been other works on VIST. Most of them extended the encoder-decoder architecture introduced in the baseline publication by adding attention mechanisms \cite{DBLP:journals/corr/abs-1805-10973}, learning positionally dependent parameters \cite{DBLP:journals/corr/abs-1806-00738} and using reinforcement learning based methods \cite{DBLP:journals/corr/abs-1804-09160}, \cite{DBLP:journals/corr/abs-1805-08191}.

To our best knowledge, there are no prior works making use of characters for visual storytelling. The only work that uses any additional semantics for story generation is \cite{DBLP:journals/corr/abs-1805-08191}. They propose a hierarchical model structure which first generates a ``semantic topic" for each image in the sequence and then uses that information during the generation phase. The core module of their hierarchical model is a Semantic Compositional Network (SCN) \cite{DBLP:journals/corr/GanGHPTGCD16}, a recurrent neural network variant generating text conditioned on the provided semantic concepts.

Unlike traditional attention mechanisms, the SCN assembles the information on semantics directly into the neural network cell. It achieves this by extending the gate and state weight matrices to adhere to additional semantic information provided for the language generation phase. Inspired by the results SCN achieved for image and video captioning, we use it for storytelling. The semantic concepts we use are based on character frequencies and their co-occurrence information extracted from the stories of the VIST dataset. 

Our expectation is that the parameters of the language decoder network generating the story are dependent on the character semantics and would learn to capture linguistic patterns while simultaneously learning mappings to respective visual features of the image sequence.

\section{Data}

We used the Visual storytelling (VIST) dataset comprising of image sequences obtained from Flickr albums and respective annotated descriptions collected through Amazon Mechanical Turk \cite{DBLP:journals/corr/HuangFMMADGHKBZ16}. Each sequence has 5 images with corresponding descriptions that together make up for a story. Furthermore, for each Flickr album there are 5 permutations of a selected set of its images. In the overall available data there are 40,071 training, 4,988 validation, and 5,050 usable testing stories.

\subsection{Character extraction}

We extracted characters out of the VIST dataset. To this end, we considered that a character is either ``a person" or ``an animal". We decided that the best way to do this would be by making use of the human-annotated text instead of images for the sake of being diverse (e.g.: detection on images would yield ``person", as opposed to father). 

The extraction takes place as a two-step process:

\textbf{Identification of nouns}: We first used a pretrained part-of-speech tagger \cite{Marcus:1994:PTA:1075812.1075835} to identify all kinds of nouns in the annotations. Specifically, these noun categories are NN -- common, singular or mass, NNS -- noun, common, plural, NNP -- noun, proper, singular, and NNPS -- noun, proper, plural.

\textbf{Filtering for hypernyms}: WordNet \cite{Miller:1995:WLD:219717.219748} is a lexical database over the English language containing various semantic relations and synonym sets. Hypernym is one such semantic relation constituting a category into which words with more specific meanings fall. From among the extracted nouns, we thereby filtered those words that have their lowest common hypernym as either ``person" or ``animal".

\subsection{Character analysis}

We analyzed the VIST dataset from the perspective of the extracted characters and observed that 20,405 training, 2,349 validation and 2,768 testing data samples have at least one character present among their stories. This is approximately 50\% of the data samples in the entire dataset.
To pursue the prominence of relationships between these characters, we analyzed these extractions for both individual and co-occurrence frequencies.

\begin{figure}[ht]
    \includegraphics[width=0.5\textwidth]{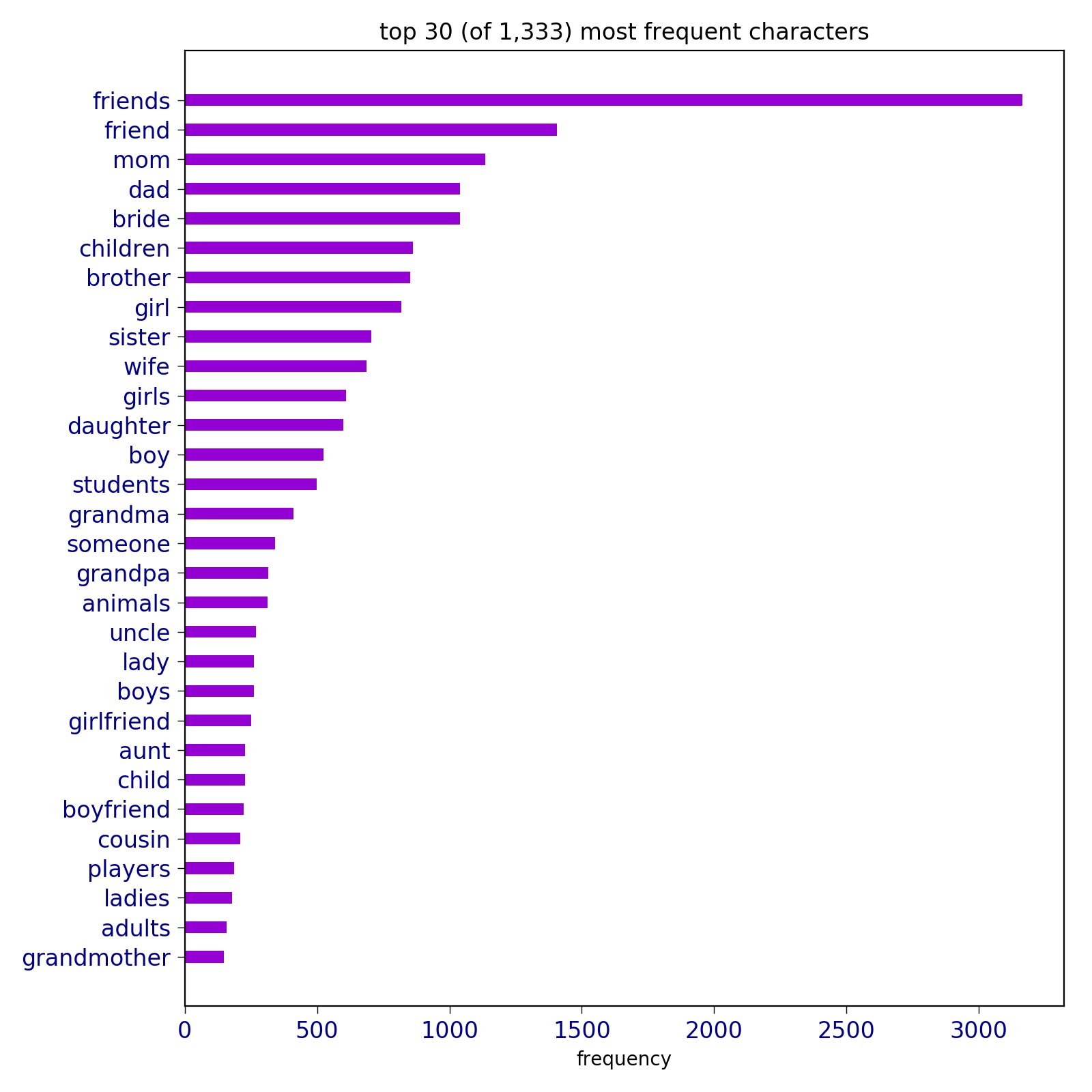}
    \caption{Character frequencies (training split).}
    \label{fig:char_freqs}
\end{figure}

\begin{figure}[ht]
    \includegraphics[width=0.5\textwidth]{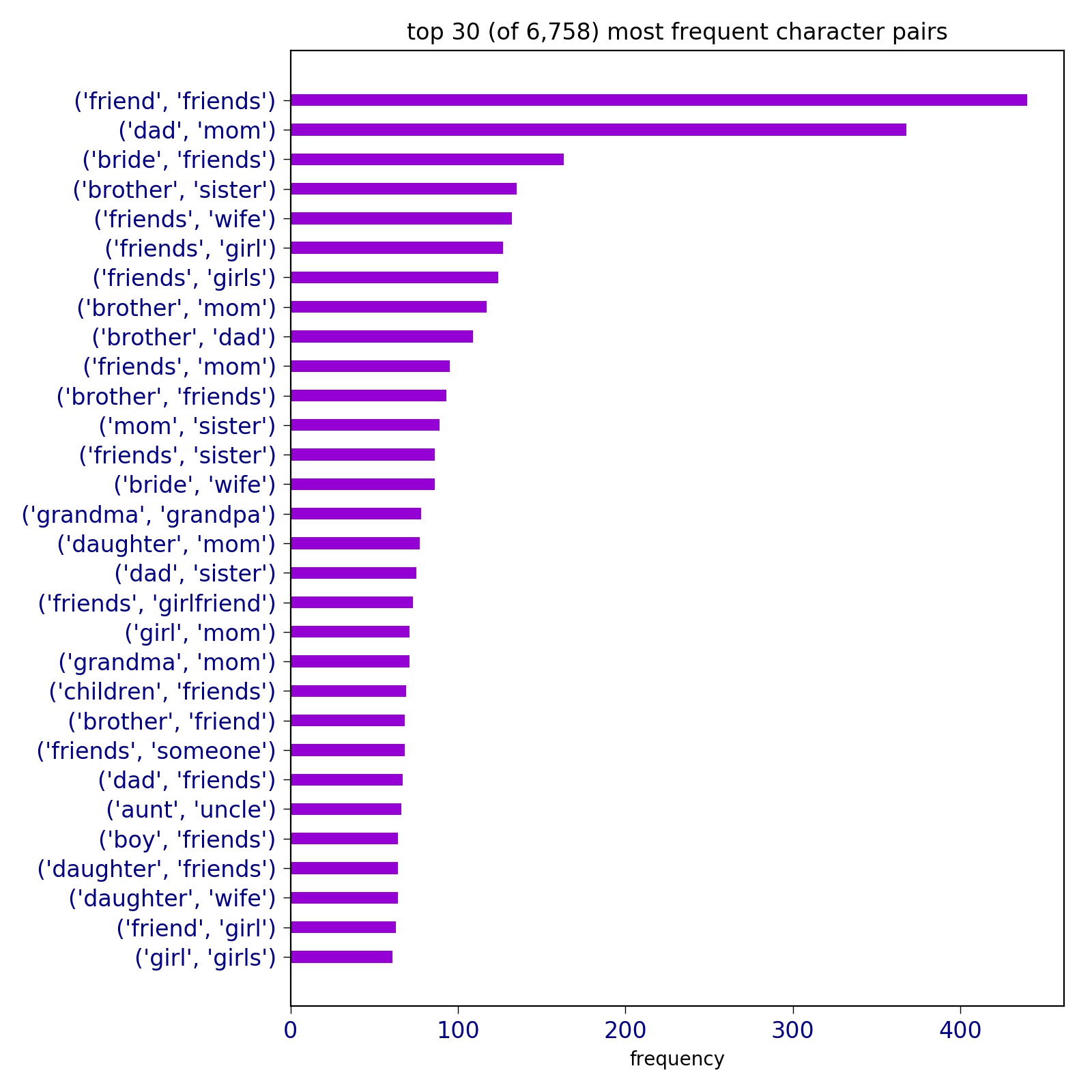}
    \caption{Characters co-occurrence frequencies (training split).}
    \label{fig:char_pair_freqs}
\end{figure}

We found a total of 1,470 distinct characters with 1,333 in training, 387 in validation and 466 in the testing splits. This can be considered as an indication to the limited size of the dataset because the number of distinct characters within each split is strongly dependent on the respective size of that split. 

Figure \ref{fig:char_freqs} plots the top 30 most frequent characters in the training split of the dataset. Apart from the character ``friends" there is a gradual decrease in the occurrence frequencies of the other characters from ``mom" to ``grandmother". Similarly, in Figure \ref{fig:char_pair_freqs}, which plots the top 30 most co-occurring character pairs, (``dad", ``mom"), (``friend", ``friends")  pairs occur drastically more number of times than other pairs in the stories. This can lead to an inclination bias of the story generator towards these characters owing to the data size limitations we discussed.

In the process of detecting characters, we observed also that $\sim$5,000 distinct words failed on WordNet due to their misspellings (``webxites"), for being proper nouns (``cathrine"), for being an abbreviation (``geez"), and simply because they were compound words (``sing-a-long"). Though most of the models ignore these words based on a vocabulary threshold value (typically 3), we would like to comment that language model creation without accounting for these words could adversely affect the behavior of narrative generation.

\section{Model}

\begin{figure}[ht]
    \includegraphics[width=0.5\textwidth]{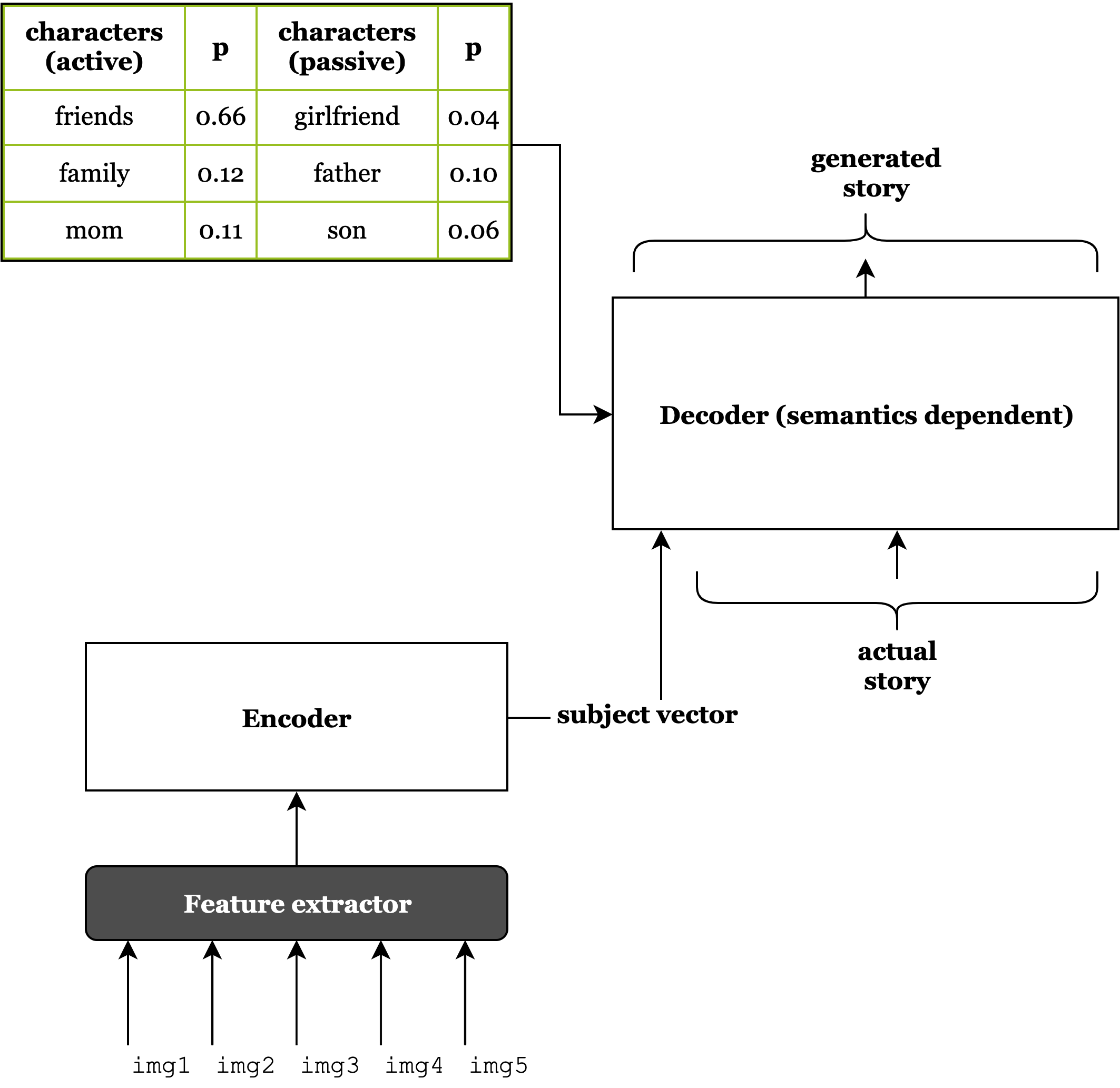}
    \caption{The model follows the encoder-decoder structure. Additional character semantics passed to the decoder module regulate its state parameters.}
    \label{fig:model}
\end{figure}

Our model in Figure \ref{fig:model} follows the encoder-decoder structure. The encoder module incorporates the image sequence features, obtained using a pretrained convolutional network, into a subject vector. The decoder module, a semantically compositional recurrent network (SCN) \cite{DBLP:journals/corr/GanGHPTGCD16}, uses the subject vector along with character probabilities and generates a relevant story.

\subsection{Character semantics}
\label{char_sem}

The relevant characters with respect to each data-sample are obtained as a preprocessing step. We denote characters extracted from the human-annotated stories of respective image-sequences as \emph{active} characters. We then use these active characters to obtain other characters which could potentially influence the narrative to be generated. We denote these as \emph{passive} characters and they can be obtained using various methods. We describe some methods we tried in Section \ref{setup}. The individual frequencies of these relevant characters, active and passive are then normalized by the vocabulary size and constitute the character probabilities.

\subsection{Encoder}

Images of a sequence are initially passed through a pretrained ResNet network \cite{DBLP:journals/corr/HeZRS15}, for obtaining their features. The features extracted are then provided to the encoder module, which is a simple recurrent neural network employed to learn parameters for incorporating the subjects in the individual feature sets into a subject vector.

\subsection{Decoder}

\begin{figure}[h!]
    \includegraphics[width=0.5\textwidth, height=0.5\textwidth]{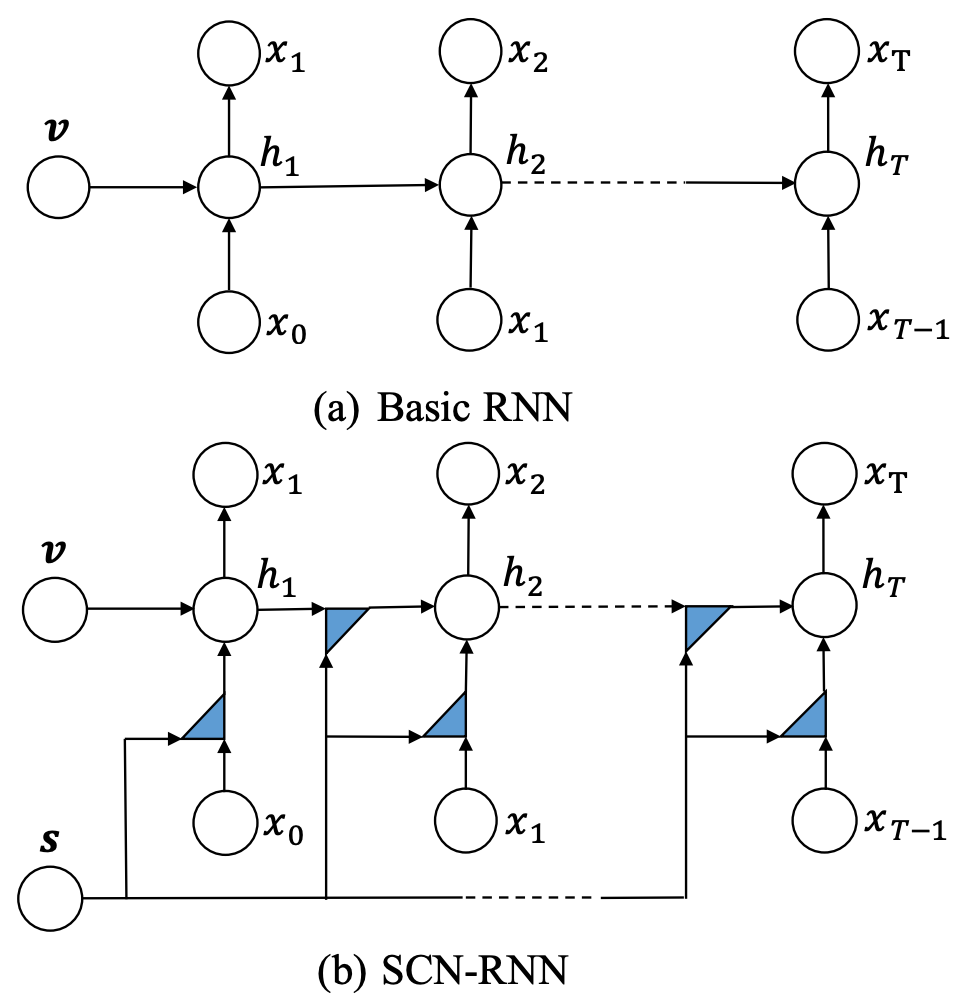}
    \caption{\cite{DBLP:journals/corr/GanGHPTGCD16}, $v$ and $s$ denote the visual and semantic features respectively. Each triangle symbol represents an ensemble of tag dependent weight matrices.}
    \label{fig:scn}
\end{figure}

We use the SCN-LSTM variant of the recurrent neural network for the decoder module as shown in Figure \ref{fig:scn}. The network extends each weight matrix of the conventional LSTM to be an ensemble of a set of tag-dependent weight matrices, subjective to the character probabilities. Subject vector from the encoder is fed into the LSTM to initialize the first step. The LSTM parameters utilized when decoding are weighted by the character probabilities, for generating a respective story.

Gradients $\nabla$, propagated back to the network, nudge the parameters $W$ to learn while adhering to respective character semantic probabilities $s$:
\begin{equation}
\label{eq:1}
\nabla(W_{\text{gates, states}} \,|\, s, v) \:=\:  \alpha \,\cdot\, \nabla_{\text{gates, states}} \;\;\; .
\end{equation}
Consequently, the encoder parameters move towards incorporating the image-sequence features better.
Another semantic-decoder we experimented with is the LSTM-RT2 \cite{DBLP:journals/corr/GanGHPTGCD16} variant which reads character semantics as supplementary input at every timestep of the module.

\section{Experiments and results}
\label{setup}

As mentioned in Section \ref{char_sem}, passive characters can be selected by conditioning their relationships on several factors. 
We explain two such methods:

\subsection{Method 1}
\label{sec:method1}

In the first method we na\"ively select all the characters co-occurring with respective active characters. Subsequently, probabilities for these passive characters are co-occurrence counts normalized by the corpus vocabulary size. This method enables the model to learn parameters on the distribution of character relationships.

\subsection{Method 2}
\label{sec:method2}

\begin{figure}[h!]
    \includegraphics[width=0.5\textwidth]{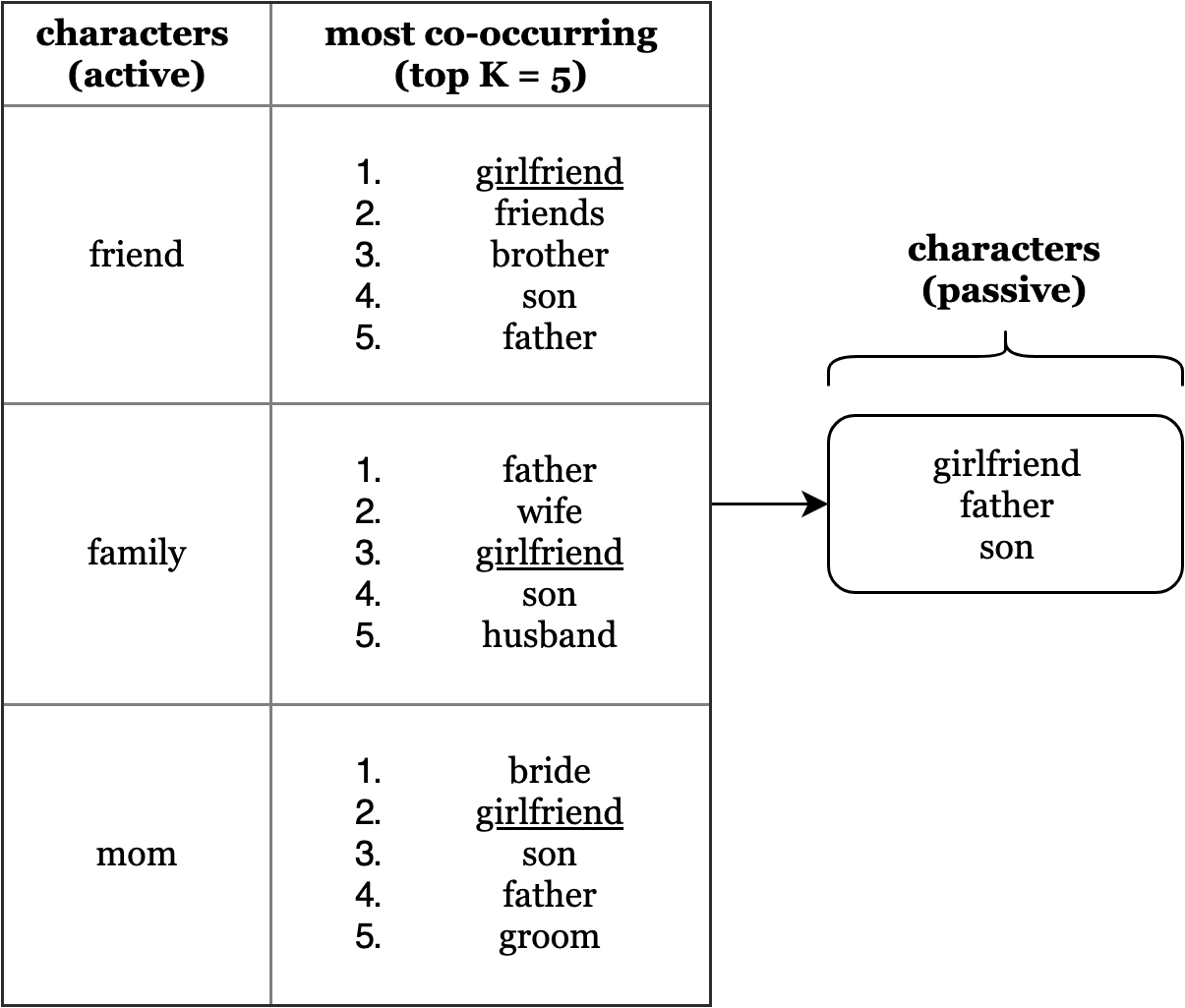}
    \caption{Conditional on collective co-occurrences.}
    \label{fig:method2}
\end{figure}

In the second approach, we conditionally select a limited number of characters that collectively co-occur most with the respective active characters. This is visualized in Figure \ref{fig:method2}. The selected passive characters ``girlfriend", ``father" and ``son" collectively co-occur in the most co-occurring characters sets of the active characters. $K$ in this case is a tunable hyperparameter. 

For the experiments, images were resized to be 224 dimensional and re-sampled using bi-linear interpolation. Upon pre-processing the images, respective features were extracted using Resnet-152 and passed through the encoder module that yields a subject vector comprising the context across the image sequence. The semantic feature vector of size 1,470 and the sequence feature vector of size 10,240 were linearly transformed to have a dimensionality of 250 for adhering to the word embedding layer dimension.

\begin{figure}
	\centering
    \captionsetup{singlelinecheck=off}
	\includegraphics[width=0.5\textwidth]{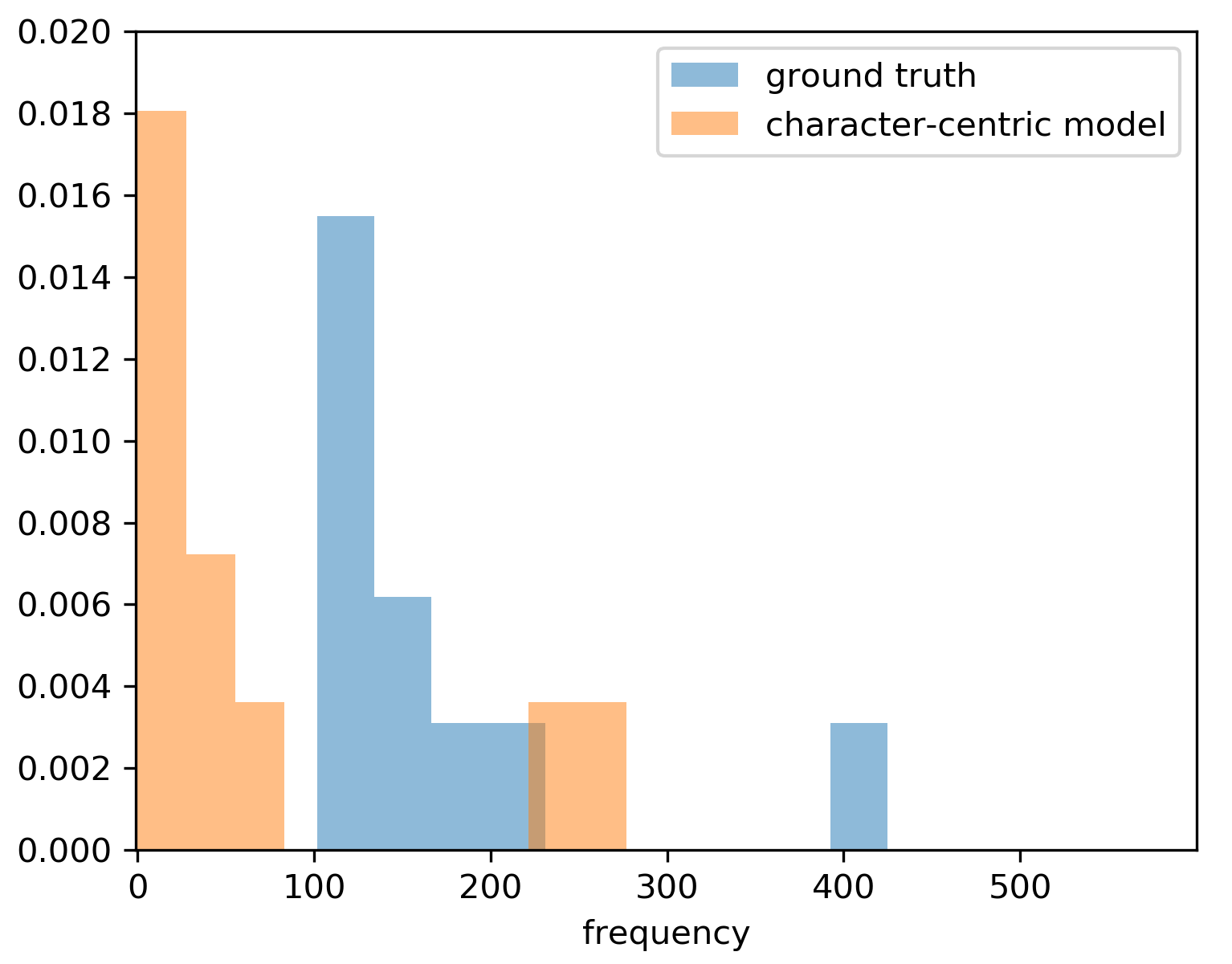}
	\caption*{Characters from the VIST test split with frequencies above 100.}
	\includegraphics[width=0.5\textwidth]{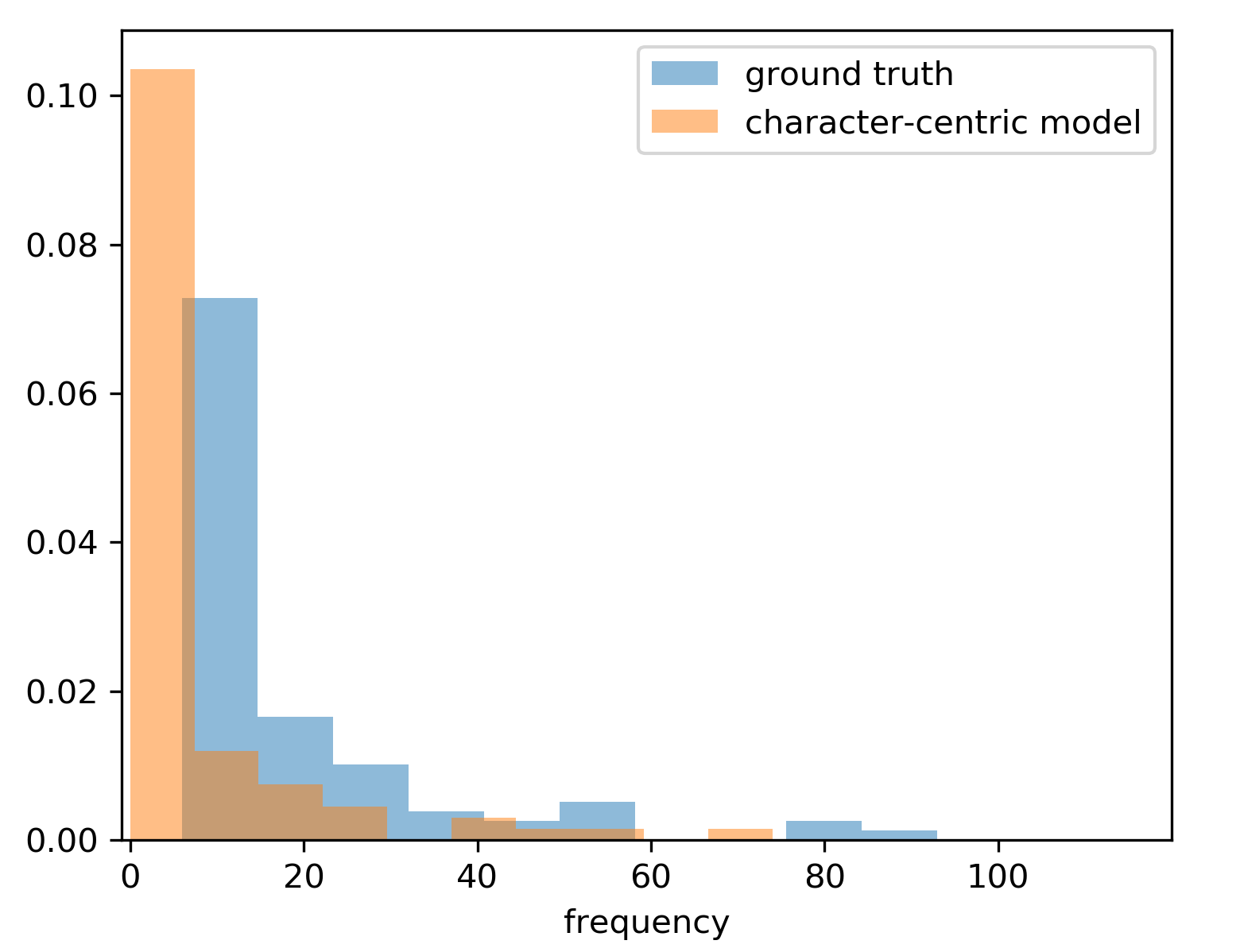}
	\caption*{Characters from the VIST test split with frequencies above 5 and below 100.}
	\includegraphics[width=0.5\textwidth]{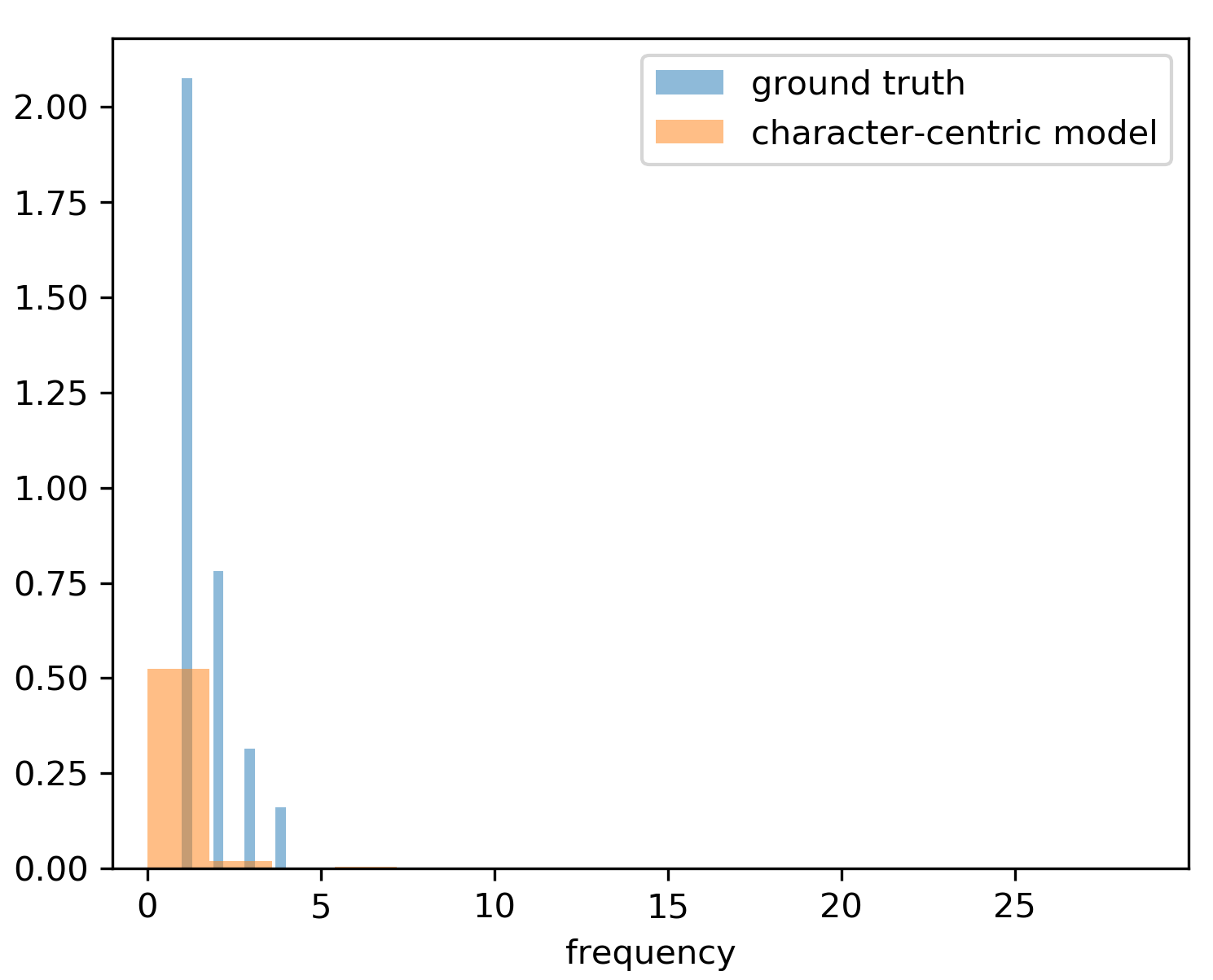}
	\caption*{Characters from the VIST test split with frequencies below 5.}
	\caption{For the purpose of analyses, the characters of the VIST test split were segregated into three tiers, i.e., those with frequencies above 100, between 5 and 100, and below 5.}
	\label{fig:cc_plots}
\end{figure}

Other details pertaining the semantic decoder include a hidden size of 1,000 for a two-layered GRU with a dropout of 0.5. Cross-entropy loss is employed as the training criterion with the Adam optimization algorithm and the learning rate of 0.0001. The model is trained for 100 epochs with a batch size of 64 and then used to generate stories for the VIST test split. The characters from the generated stories are then plotted against the true characters’ distributions of the VIST test data split, to understand the influence of the trained model. The plots are visualized in Figure \ref{fig:cc_plots} and sample stories generated by the model are provided under Appendices.

\section{Discussion}

We started off with the intention to learn the underlying spread of characters and generate them in the stories with relevancy and creativity. The frequency distributions plotted in Figure 5.6 depict that the character-centric model is learning to match the true distributions. From among the three tiers of characters in the VIST test data split, i.e, those with frequencies above 100 (most frequent), between 5 and 100, and below 5 (least frequent), distributions of the most and least frequent tiers are moving towards the ground truth.

However, in some of the stories generated, we observe that the model exaggerates and over-contextualizes the narratives with imaginary characters. This behavior can be due to the static character semantics vector used in the decoder module, without any fine-tuning. Both methods described in Sections \ref{sec:method1} and \ref{sec:method2} exhibit different traits. Empirical observations show that in the case of the VIST dataset, which is sparse with characters, the approach of selecting all co-occurrences as passive, works better. We are further working towards alternate evaluations of the character relationships learned by the models and understanding the abstract concepts that get generated as a result of such learning.

\section{Conclusion}

We have extracted, analyzed and exploited characters in the realm of storytelling using the VIST dataset. We have provided a model that can make use of the extracted characters to learn their relationships and thereby generate grounded and subjective narratives for respective image sequences. Learning behaviors, inference stage performances, and subjective aspects of the generated results are comprehensively discussed. For future work we would like to make the encoder semantically compositional by extracting visual tags and also explore ways to improve learning of character relationships while avoiding the over-fitting bottleneck. Additionally, we intend to pursue ways to analyze the actual influence of visual and textual modalities on the model outcomes for a thorough understanding of the visual storytelling paradigm.

\newpage
\bibliography{character_relationships}
\bibliographystyle{acl_natbib}

\appendix

\section{Appendices}
\label{sec:appendix}

Stories generated for sample VIST test split image sequences: 

\begin{figure*}[htbp]
	\centering
    \captionsetup{singlelinecheck=off}
    \fbox{
    \begin{minipage}{\textwidth}
    \includegraphics[width=\textwidth, height=\textheight, keepaspectratio]{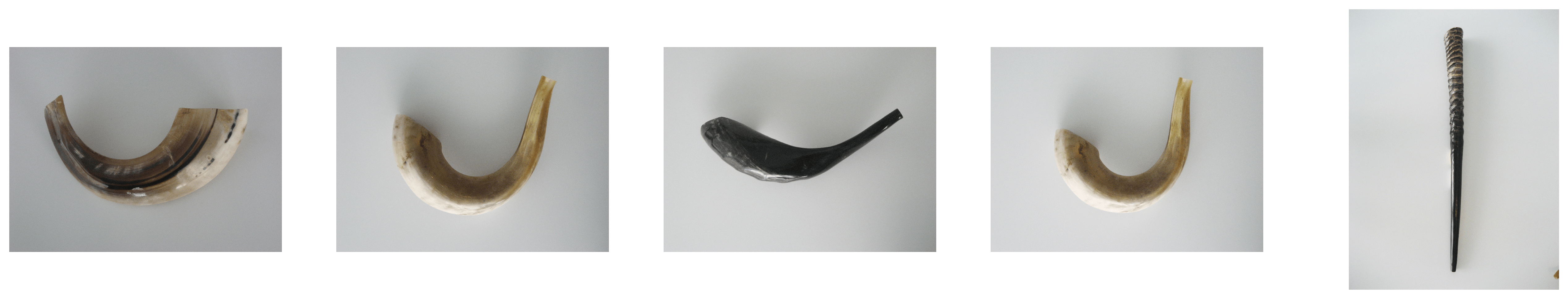}
	\caption*{\underline{Story (Character-centric model):} \emph{i went to the museum to see the dolphins at the museum . i saw some really interesting coral . there were some amazing art installations . i also saw some penguins . i also saw some dinosaurs .}\newline\newline\underline{Story (human annotated):} \emph{there are many types of horns and bone . this is a ram horn and it has been shined well . next there is a horn that has been made into a pipe . a curved horn has been rubbed and shined . finally there is a straight piece of bone that has been shaped .\newline}}
	\includegraphics[width=\textwidth, height=\textheight, keepaspectratio]{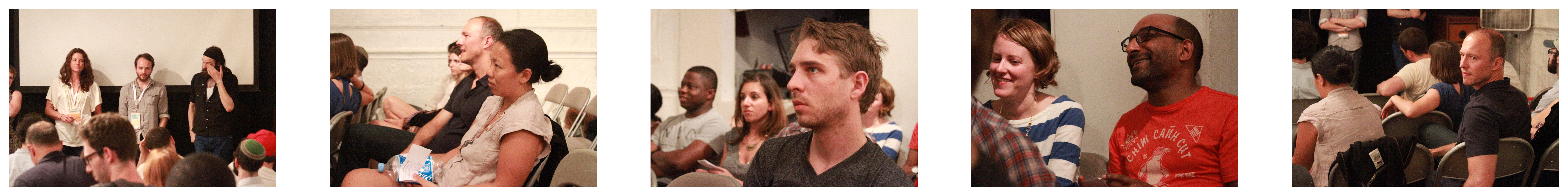}
	\caption*{\underline{Story (Character-centric model):} \emph{the crowd gathered for the awards ceremony . the speaker gave a great speech . the director gave a brief speech . the audience listened to the speaker . the cameraman was a little shy after the speech ended  .}\newline\newline\underline{Story (human annotated):} \emph{everyone from the town gathered to attend the homeowner 's association meeting . [female] did n't agree with some of the proposals raised at the meeting . [male] was just thinking about going home to eat some pizza and watch tv . the presenters made a joke about pooper scoopers . [male] looked at his wife who arrived late to the meeting .}}
	\end{minipage}
	}
	\caption{VIST test split image sequences with stories generated by the character-centric storytelling model.}
	\label{fig:ccs_results}
\end{figure*}

\end{document}